# Development and Testing of Retrieval Augmented Generation in Large Language Models - A Case Study Report

Yu He Ke*1,2
Liyuan Jin*3,4,5
Kabilan Elangovan4,5
Hairil Rizal Abdullah1,2
Nan Liu3
Alex Tiong Heng Sia3,7
Chai Rick Soh1,3
Joshua Yi Min Tung2,6
Jasmine Chiat Ling Ong3,8
Daniel Shu Wei Ting+3,4,5

Affiliations:
1 Department of Anesthesiology, Singapore General Hospital, Singapore, Singapore
2 Data Science and Artificial Intelligence Lab, Singapore General Hospital, Singapore
3 Duke-NUS Medical School, Singapore, Singapore
4 Singapore National Eye Centre, Singapore Eye Research Institute, Singapore, Singapore
5 Singapore Health Services, Artificial Intelligence Office, Singapore
6 Department of Urology, Singapore General Hospital, Singapore
7 Department of Women's Anaesthesia, KK Women's and Children's Hospital, Singapore.
8 Division of Pharmacy, Singapore General Hospital, Singapore

*Contributed Equally
+Corresponding Author

Corresponding Author:
Name: Daniel Shu Wei Ting
Email: daniel.ting45@gmail.com
Address: 31 Third Hospital Ave. Singapore 168753
Institution: Singapore Health Services, Artificial Intelligence Office, Singapore

Keywords: Large language model, artificial intelligence, Retrieval-augmented generation

Declaration:
Ethics approval and consent to participate: Not applicable
Availability of data and material: Yes

Financial Disclosure
The authors have not declared a specific grant for this research from any funding agency in the public, commercial, or not-for-profit sectors.

Competing interests
None declared.

Patient and public involvement
Patients and/or the public were not involved in the design, conduct, reporting, or dissemination plans of this research.

Acknowledgments
The authors extend their sincere gratitude to the four junior doctors from Singapore General Hospital for their invaluable contributions as human evaluators in this study. Their dedication and insightful inputs were instrumental in enriching the research. We also acknowledge the preoperative clinical guidelines provided by the hospital, which were pivotal in the successful execution of this project.

Author contributions
This study was conceptualized by YH Ke, HR Abdullah, and DSW Ting, who played pivotal roles in establishing the research framework and guiding the project's direction. The coding and technical development of the project were led by L Jin and K Elangovan, whose expertise in software engineering was critical to the implementation of the LLM-RAG models. YH Ke, L Jin, JYM Tung, JCL Ong, ATH Sia, CR Soh, MLLJames, and N Liu made contributions to the analysis of data and drafting of the manuscript.

**Abstract**

**Purpose:** Large Language Models (LLMs) hold significant promise for medical applications. Yet, their practical implementation often falls short in incorporating current, guideline-grounded knowledge specific to clinical specialties and tasks. Additionally, conventional accuracy-enhancing methods like fine-tuning pose considerable computational challenges.

Retrieval Augmented Generation (RAG) emerges as a promising approach for customizing domain knowledge in LLMs, particularly well-suited for needs in healthcare implementations. This case study presents the development and evaluation of an LLM-RAG pipeline tailored for healthcare, focusing specifically on preoperative medicine. The accuracy and safety of the responses generated by the LLM-RAG system were evaluated as primary endpoints.

**Methods:** We developed an LLM-RAG model using 35 preoperative guidelines and tested it against human-generated responses, with a total of 1260 responses evaluated (336 human-generated, 336 LLM-generated, and 588 LLM-RAG-generated).

The RAG process involved converting clinical documents into text using Python-based frameworks like LangChain and Llamaindex, and processing these texts into chunks for embedding and retrieval. Vector storage techniques and selected embedding models to optimize data retrieval, using Pinecone for vector storage with a dimensionality of 1536 and cosine similarity for loss metrics. LLMs including GPT3.5, GPT4.0, Llama2-7B, llama2-13B, and their LLM-RAG counterparts were evaluated.

We evaluated the system using 14 de-identified clinical scenarios, focusing on six key aspects of preoperative instructions. The correctness of the responses was determined based on established guidelines and expert panel review. Human-generated answers, provided by junior doctors, were used as a comparison. Comparative analysis was conducted using Cohen's H and chi-square tests.

**Results**: The LLM-RAG model generated answers within an average time of 15-20 seconds, significantly faster than the 10 minutes typically required by humans. Among the basic LLMs, GPT4.0 exhibited the best accuracy of 80.1%. This accuracy was further increased to 91.4% when the model was enhanced with RAG. Compared to the human-generated instructions, which had an accuracy of 86.3%, the performance of the GPT4.0-RAG model demonstrated non-inferiority (p=0.610).

**Conclusions:** In this case study, we demonstrated a LLM-RAG model for healthcare implementation. The model can generate complex preoperative instructions across different clinical tasks with accuracy non-inferior to humans and low rates of hallucination. The pipeline shows the advantages of grounded knowledge, upgradability, and scalability as important aspects of healthcare LLM deployment.

Graphical abstract:

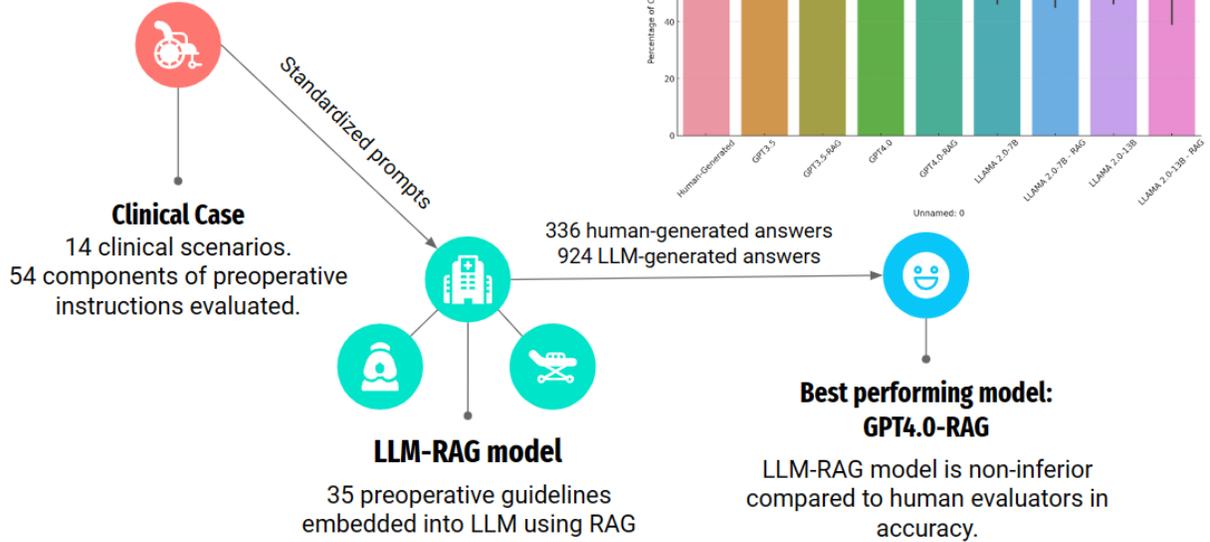

## Introduction

Large Language Models (LLMs) have gained significant attention for their clinical applications potential[1,2], and have been demonstrated to match human performance in basic clinical tasks such as rating American Society of Anesthesiologists (ASA) scoring[3]. However, where complex tasks, such as clinical assessment and management are given, the response only relies on pre-train knowledge and is not grounded on institutional practicing guidelines. Most importantly, hallucinations from LLMs pose significant safety and ethical concerns[4,5].

Surgery cancellations on the day of surgery due to medical unfitness[6], incorrect physician instructions[7], and non-compliance to preoperative instructions[8] pose a significant economic impact[9], with operating room expenses estimated between USD 1400 to 1700 per hour[10]. Thorough preoperative evaluations can minimize these cancellations[11], but traditional preoperative evaluations are inherently labor-intensive and costly. The utilization of domain-specific LLM for delivering preoperative instructions presents substantial potential for personalized preoperative medicine.

**Finetuning vs Retrieval Augmented Generation (RAG)**
Fine-tuning, a process involving the retraining of pre-existing foundational models with relatively small datasets and a limited number of epochs, presents a viable solution to mitigate the issue of hallucination in model outputs. While this approach does lead to a reduction in the generalizability of the pre-trained model, as it becomes more specialized for a specific task, it offers several advantages. These include diminished instances of hallucination, enhanced performance tailored to the specific task, and increased consistency in response generation[12].

The primary challenges in fine-tuning LLMs stem from various factors including the need for extensive retraining datasets; the rigidity of question-and-answer format for both input and output; the substantial time investment required to generate accurate answers, particularly for complex fields like healthcare; and technical hurdles such as limitations in context tokens and the computational demands typically quantified in petaflops for GPU memory[13,14].

Retrieval Augmented Generation (RAG) is an innovative approach for tailoring LLMs to specific tasks. Unlike traditional LLMs, RAG functions similarly to a search engine, retrieving relevant, customized text data in response to queries. This capability effectively turns RAG into a tool that integrates specialized knowledge into LLMs, enhancing their baseline capabilities. In healthcare, for instance, LLMs equipped with RAG and embedded with extensive clinical guidelines (LLM-RAG) can yield more accurate outputs[15]. Currently, two primary open-source frameworks for RAG exist - LangChain[16] and Llamaindex[17]. Although the retrieval process of RAG can be technically challenging, RAG's utility in contexts with smaller, more focused knowledge corpora remains significant. A detailed comparison of fine-tuning versus RAG is presented in Table 1.

**Table 1.** A detailed comparison between finetuning and RAG*.

| Methods | Main Feature | Pros | Cons |
|---------|--------------|------|------|

| Finetuning | A customized LLM retrained for any intended task. | Consistent response generation for specific tasks. Cost efficient in deployment (cloud). | Computationally costly during the finetuning stage. Requires large finetuning datasets (100-500 Q/A pairs). Requires datasets to be curated by clinical professionals. The response could be narrow and rigid. |
|---|---|---|---|
| **RAG** | A retrieval framework to provide pre-trained LLMs with customized knowledge. | Easy to update Answering to various queries. Maintains reasoning ability of pre-trained models. | A large knowledge corpus introduces noise. Does not work well with highly structured data. |

*Depends on the size of selected LLM models (or pre-trained models for RAG)

This study aims to detail the development of a LLM-RAG pipeline, tailored specifically for preoperative medicine. The primary objective is to evaluate the accuracy of the LLM-RAG in determining a patient's fitness for surgery. The secondary objective involves a comprehensive assessment of the LLM-RAG's accuracy and safety in providing various preoperative instructions.

## Methods

### Preoperative Guidelines
A comprehensive collection of 35 preoperative guidelines with detailed protocols for patient assessment, medication management, and specific surgery were used (Supplementary Table 1). These documents are adopted from international perioperative guidelines[18–21]. These protocols are extracted in their original PDF format including diagrams and figures.

### Retrieval-Augmented Generation (RAG) Framework
The LLM-RAG pipeline framework is composed of multiple distinct components:

### RAG:
### A. Preprocessing
Clinical documents can come in various formats. To make these compatible with the RAG framework, they should be converted into a text-based format. Open-source, Python-based frameworks like Langchain and Llamaindex facilitate this conversion, providing loaders for each document type. These loaders not only transform the documents into text but also preserve additional information in metadata, enhancing the document retrieval process by allowing for the screening of titles and pages. However, this automated conversion process has limitations. Key details, such as diagrams, decision trees, and algorithms may not be interpreted appropriately and lead to an increase in noise.

The processed texts are then segmented into smaller chunks for embedding and retrieval. Tools from Huggingface and Langchain assist in this splitting process, typically creating chunks of arbitrary size with slight overlaps. Determining the ideal chunk size for healthcare applications is challenging and requires qualitative assessment. Advanced RAG techniques developed by Llamaindex, such as the automated parameter search, Sentence Window retrieval, and Auto-merging retrieval, show promise in addressing these challenges[22].

In our case study, we utilized Python 3.11 with Langchain's DirectoryLoader and RecursiveCharacterTextSplitter, and the PyPDF2 package for reading PDF-based clinical guidelines. The chunk size was set at 1000 units with a 100-unit overlap, based on the cl100k_base tokenizer and the selected embedding model.

### B. Vector Storage and Embedding Model

Optimizing data retrieval necessitates data storage engineering. Vector Storage, based on similar ideation in deep learning techniques, efficiently condenses information into high-dimensional vectors, enhancing retrieval performance. Key metrics for evaluating Vector Storage include cosine similarity, Euclidean distance, and dot product. Cosine similarity, particularly effective for semantic searches, is generally preferred for healthcare applications due to its focus on the angle between vectors, thereby emphasizing content similarity. In contrast, Euclidean distance is more suitable for quantitative measurements.

Embedding, the process of converting content into a numerical representation (i.e., vectors) for machine learning models, is crucial for transforming preprocessed healthcare knowledge into individual vectors. There are several embedding models available for this purpose. Cloud-based options like OpenAI's text-embedding-ada-002 and local options such as Hugging Face's all-MiniLM-L6-v2 offer varied indexing and dimensions. It is essential to align the Vector Storage parameters with the selected embedding model's specifications.

In this case study, we used Pinecone as our cloud-based Vector Storage solution, configured with 1536 dimensions and cosine similarity as the loss metric. For embedding, we utilized OpenAI's text-embedding-ada-002 model.

### C. Retrieval Agent

The Retrieval Agent functions as an intermediary, pinpointing the most relevant knowledge chunks in response to user queries. This process involves using the same embedding model to convert the text from the query into a vector. It then sifts through the Vector Storage to identify the closest matching vectors. Additionally, the agent can access pre-processed metadata for further evaluation. The number of chunks retrieved, adjustable through a parameter commonly referred to as 'k', can be set to any desired value. The effectiveness of a Retrieval Agent is closely tied to the underlying framework it is built upon. Each framework, be it Langchain with Pinecone, Chroma, or Llamaindex, requires its unique Retrieval Agent to function optimally.

In this case study, we utilized Pinecone's Retrieval Agent, setting 'k' to 10 to determine the number of knowledge chunks to retrieve. Both the chunk size and k is selected with Llamaindex scoring among various combination with a top score of 0.84 on average retrieval, on another independent evaluation dataset.

**Large Language Model and Response Generation**

A list of pre-trained foundational LLMs is selected for this case study, including GPT 3.5[23], GPT 4.0[24], LLAMA2-7B[25,26], and LLAMA2-13B[26]. Detailed characteristics of these LLMs are provided in Table 3. The same set of knowledge extracted from the guideline knowledge corpus was used as user prompts in various scenarios. Additionally, the clinical questions were input as system prompts for all models.

For LLM inference, the response generation process was conducted using different platforms depending on the model. GPT-based models were run on the cloud-based OpenAI playground platform, while the LLAMA2 models utilized the open-source H2O GPT platform[27]. Notably, we employed quantized 8-bit chat models for both LLAMA2 7B and LLAMA2 13B models, operating on a local Nvidia RTX 4090 GPU.

To strike a balance between minimizing hallucinations and maintaining rigorous generation, the temperature settings for all LLMs were set at 0 for GPT models and 0.1 for LLAMA2 models. The detailed RAG framework can be found in Table 2. For complete transparency and to facilitate further research, the entire codebase has been made publicly available on GitHub at [RetinAIs/RAG-LLM-Demo (github.com)](.).

**Table 2:** Detailed RAG framework components and available tools.

|  | **Preprocessing** | **Text Splitter** | **Embedding Model** | **Vector Storage** | **Retrieval Agent** |
|---|---|---|---|---|---|
| Function | Help convert any healthcare document format into a text-based | Help make processed text-based documents into smaller chunks | Help covert chunks into their vectors | A storing space for all converted vectors | A proxy to retrieve the most relevant documents |
| Common Tools | Text Convert (Python):<br>Word: Docx2txt<br>PDF: PyPDF2<br>CSV: Pandas<br><br>Loder:<br>Langchain:<br>Text: TextLoader<br>CSV:: CSVLoader<br>Directory: DirectoryLoader<br>PDF: | Langchain: RecursiveCharacterTextSplitter<br><br>Llamaindex: SentenceSplitter | OpenAI (Cloud): Text-embedding-ada-002<br><br>Hugging Face (Local): all-MiniLM-L6-v2 | Cloud: Pinecone<br><br>Local: Chroma Llamaindex | Pinecone: Pinecone.index<br><br>Chroma: Similarity_search<br><br>Llamaindex: retriever |

|  | PyPDFLoader Llamaindex: CSV: SimpleCSVReader PDF: PDFReader Docx: DocxReader |  |  |  |  |

Detailed in Figure 1 is the operational framework of the LLM-RAG model, providing a schematic representation of the interplay of the algorithmic workflow integral to the system's functionality.

**Figure 1:** Operational framework of the LLM-RAG model incorporating 35 preoperative guidelines.

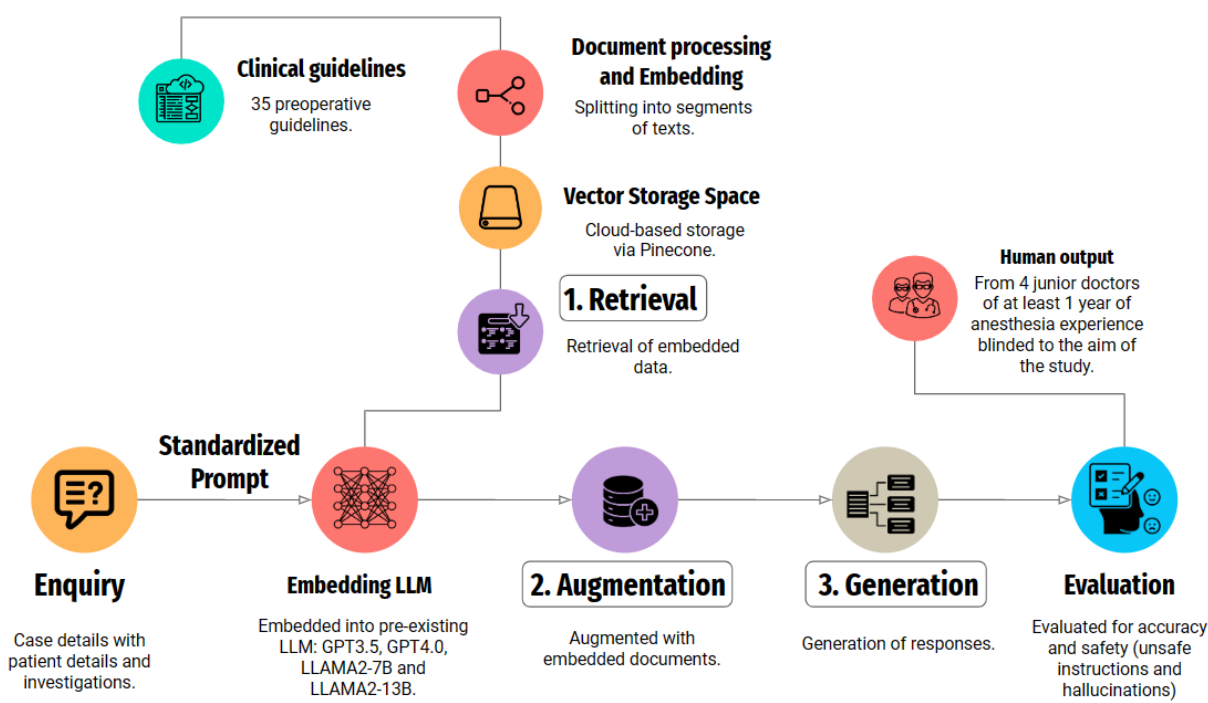

**Table 3:** Comparison of the different large language models used in the study.

| Model | Training Corpus | Model Size | Data Size | Other Key Features | Temperature chosen |
|---|---|---|---|---|---|
| GPT-3.5 | Internet text (up to 2021) | 175 billion parameters | 45 TB of text | Advanced text generation and understanding | 0 |
| GPT-4.0 | Diverse internet sources (up to 2022) | Larger than 175 billion parameters | Estimated to be larger than 45 TB | Enhanced text generation, understanding, and multimodal capabilities | 0 |
| LLAMA | Varied | 7 billion | Not publicly | Focused on specific tasks | 0.1 |
| 2-7B | (details not publicly disclosed) | parameters | disclosed | and domains | |
| LLAMA 2-13B | Varied (details not publicly disclosed) | 13 billion parameters | Not publicly disclosed | Focused on specific tasks and domains with a larger capacity | 0.1 |

**Clinical scenarios**

This study evaluated the LLM-RAG system using 14 de-identified clinical scenarios, covering a wide range of patients and surgical complexities. Six key aspects of preoperative instructions were assessed: fasting guidelines, preoperative carbohydrate loading, medication instructions, healthcare team directives, required preoperative optimizations, and the necessity to delay surgery for further optimization. These aspects were chosen due to their recognized importance in existing medical literature[28]. An example is detailed in Supplementary Table 2.

**Output**

Human-generated answers were provided by four junior doctors with 1-5 years of experience in anesthesia. They were familiar with the guidelines through their work in preoperative clinics. They are blinded to the study's aims to ensure unbiased, natural responses.

The 'correct' answers in the study were based on established preoperative guidelines and reviewed by an expert panel. In ambiguous cases, like the suspension of ACE inhibitors before surgery, both potential answers were considered correct. The study focused on scenarios with clear directives regarding the postponement of operations for additional medical workup.

Evaluation criteria revolved around accuracy and safety. Responses with at least 75% accuracy in instructions were deemed "correct". However, any response containing a significant medical error or hallucination was categorized as "wrong (hallucination)".

A comparative analysis is performed against the human-generated responses and the best-performing LLM-RAG model using Cohen's H and chi-square test. All statistical evaluations are performed in the Python 3.6 environment.

## Results

A total of 1260 components were evaluated (336 human-generated and 924 LLM-generated). The GPT4.0-RAG model emerged as the most accurate, demonstrating a non-inferior performance (307/336 (91.4%)) compared to human-generated answers (290/336 (86.3%)) (Figure 2). The LLM-RAG models took on average 1 second for retrieval and 15-20 seconds for results generation, while the human evaluators took an average of 10 minutes to generate the preoperative instructions.

GPT4.0-RAG particularly excelled in categories where human evaluators showed lower accuracy. For instance, in the category of Preoperative Carbohydrate Loading, GPT4.0-RAG achieved an 82.1%(46/56) accuracy rate compared to human-generated answers(42/56 (75.0%)) (Supplementary Table 3). Similarly, in identifying the need for preoperative optimization, GPT4.0-RAG surpassed the human score by a notable margin (47/56 (83.9%) vs 38/56 (67.9%)). The second best-performing LLM model is the GPT4.0 model without the RAG embedding. Both models have similar hallucination rates of 1.2%.

There were small overall differences in performance when comparing human-generated and GPT4.0-RAG answers. However, there was a moderate difference in answers provided in the categories of Preoperative Optimization Required (Cohen's H = - 0.38), and Need to Delay the Operation (Cohen's H = - 0.61) with better performance by GPT4.0-RAG (Supplementary Table 4).

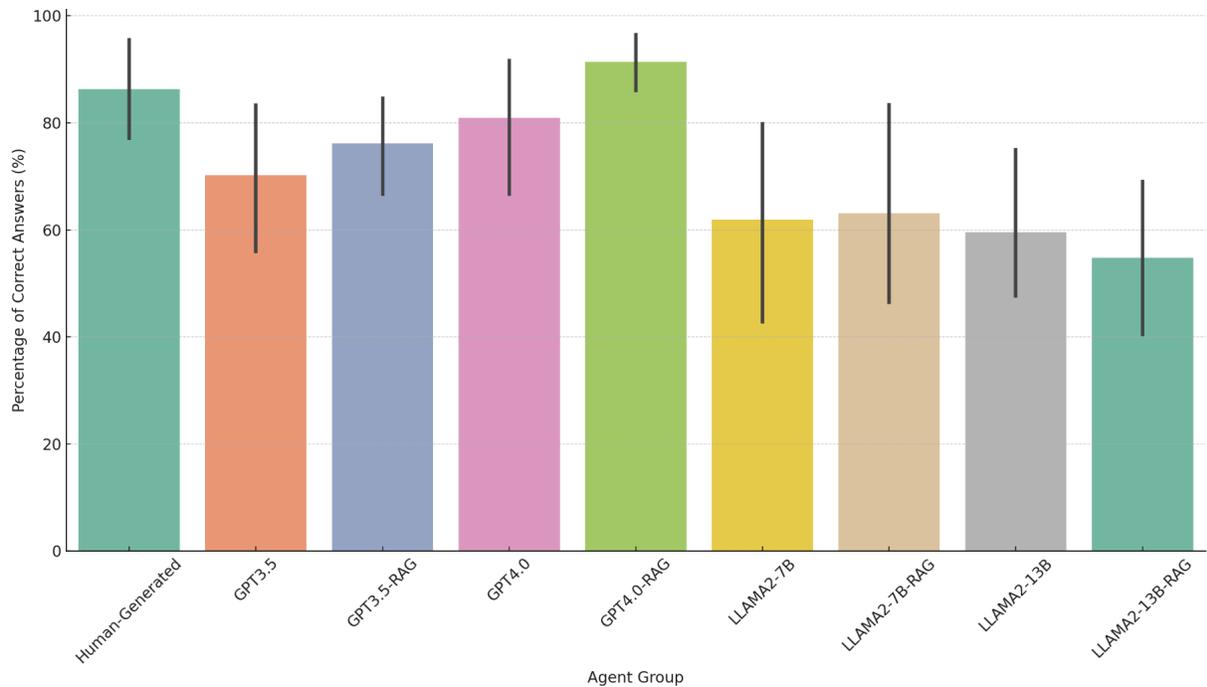

**Figure 2:** Percentage of accurate answers across the different groups.

## Discussion

We illustrated the feasibility of an LLM-RAG model for a subspecialty domain in healthcare, in this case, tailored to preoperative medicine. Our findings revealed that the accuracy of these LLM-RAG models is comparable to human evaluators.

### LLM and Domain-specific Models

The capability of LLM-RAG models to process vast amounts of data and generate responses based on comprehensive, updated guidelines positions them as potentially valuable tools in standardizing preoperative assessments. Furthermore, the emergence of fine-tuned models with ULMS[29] and BioMedLM from Stanford-CRFM[30] exemplifies the trend toward specialization in LLM applications. These domain-specific models are tailored to understand and process medical information, offering enhanced accuracy and relevance in clinical settings[31].

### LLM-RAG as a subspecialty clinical aid

The results of our study are particularly relevant in the context of the evolving landscape of elective surgical services, which have increasingly shifted towards day surgery models, reduced hospital stays, and preoperative assessments conducted in outpatient clinics[32]. The potential role of LLM-RAG in this setting as a clinical adjunct is, therefore, of considerable interest as manpower constraints span across medical providers.

Furthermore, subjectivity in clinical decisions due to variations in human judgment underscores the potential value of LLM-RAG systems in enhancing consistency in clinical decision-making. GPT models, for example, have demonstrated more consistent responses compared to anesthesiologists in tasks like ASA scoring[3]. This consistency is a crucial advantage, particularly in a field where uniformity in evaluation and decision-making can significantly impact patient outcomes.

### LLM-RAG and environmental sustainability

The adoption of LLM-RAG models may also offer benefits in environmental sustainability, particularly when compared to fine-tuning, which requires large computation power[13,14]. This process can be resource-intensive, contributing to a higher carbon footprint[33]. In contrast, LLM-RAG models allow for efficient access to domain-specific information without the need for extensive retraining. The cost of building an LLM-RAG model could be further brought down with the latest GPT4 preview, offering a lower cost per token and a much larger context size of 128k (vs 8k for GPT 4.0). It is worth comparing the performance of RAG between GPT 4.0 and GPT 4 preview, which still needs to be done in this report. Furthermore, an objective assessment of RAG retrieval contents might add another insight for better LLM-RAG model design.

### Challenges and Limitations

The study's findings are based on simulated clinical scenarios, and the LLM-RAG guidelines are based on a specific set of 35 preoperative hospital guidelines, which may limit the generalizability.

While the rate of hallucinations in the GPT4.0 models is low, the occurrence of factually incorrect or misleading data can pose significant risks. The integration of AI in healthcare must be approached with caution, ensuring that these systems are used to complement, not replace, human expertise. Additionally, the field of perioperative science is continuously evolving. The LLM-RAG models would require regular updates to keep abreast of the latest medical guidelines and evidence-based medicine. This underscores the need for continuous training and updating of these models. Furthermore, a benchmarked evaluation framework for RAG-LLM models in medicine is still lacking, underscoring the need for a cautious approach in clinical implementation[34].

The ethical implications and inherent biases of employing LLMs in clinical environments demand careful consideration. In this study, the chosen clinical scenarios were structured to yield clear decisions about delaying surgeries for medical optimization. However, real-world clinical situations often involve nuanced decisions, particularly in critical areas like cancer treatment, where the choice to postpone surgery exists in a realm of ethical ambiguity.

Users of LLM-RAG models must recognize that in complex ethical landscapes where nuanced recommendations are needed, the model might lean towards certain decisions influenced by its training data. These models are best utilized as supportive tools that complement but do not replace, the expert judgment of medical professionals.

**Conclusion**
The LLM-RAG model using GPT4.0 demonstrates significant promise in generating accurate and safe preoperative instructions and is non-inferior compared to junior doctors. This study underscores the potential role of LLM in augmenting preoperative medicine. The integration should be approached as a complementary tool to human expertise, necessitating ongoing oversight and careful implementation.

**SUPPLEMENTARY**

**Supplementary Table 1:** List of preoperative guidelines and details of the guidelines.

|  | Guideline Name | Details |
|---|---|---|
| 1 | Enhanced Recovery After Surgery for Laparoscopic Radical Nephrectomy Surgery | ERAS protocol for laparoscopic nephrectomy. |
| 2 | TOSP Surgery Risk | Surgical Risk as listed by the table code. |
| 3 | Guidelines on Preoperative Spine Clearance for patients going for non-spine surgeries | Guideline for cervical spine clearance |
| 4 | Preoperative assessment of Respiratory Disease presenting for elective surgery | Guideline for patients with respiratory disease |
| 5 | Guidelines on Preoperative Assessment and optimization of patients with thyroid disease | For patients with thyroid disease. |
| 6 | Perioperative Management of Electrolyte Abnormalities | Guideline for deranged electrolytes |
| 7 | Preoperative investigation guidelines for patients presenting for elective surgery | List of investigations that should be ordered for patients when coming in for elective operation. |
| 8 | Guidelines on Perioperative Management of Anticoagulant and Antiplatelet Therapy | Guideline for patients on anticoagulant and antiplatelets |
| 9 | Guidelines on Preoperative Assessment of Patients with Chronic Kidney Disease | For patients with Chronic Kidney Disease |
| 10 | Prevention of contrast-induced acute kidney injury in Vascular patients undergoing angioplasty | Prevention of contrast-induced acute kidney injury in vascular patients undergoing angioplasty |
| 11 | Preoperative Cardiac Evaluation and Cardiology Referral Guide | Provide general guidance on preoperative cardiac evaluation. |
| 12 | Advanced Practice Nurse Obtaining Anaesthesia Consent in Preoperative Evaluation Clinic | Guideline for when a patient can be seen by an advanced practice nurse |
| 13 | Guideline on Perioperative Management of Patients Who Refuse Blood Transfusion | Guidelines for patients who refuse blood transfusions |

| 14 | Perioperative Guideline for Patients with History of TIA/Stroke | For patients with TIA/Stroke. |
| --- | --- | --- |
| 15 | Guidelines on Preoperative Assessment of Patients with Obstructive Sleep Apnoea | For patients with OSA. |
| 16 | Guidelines on Preoperative Assessment of Obese Patients | For patients with obesity. |
| 17 | Guidelines on Preoperative Assessment and Optimization of Patients with Hypertension | For patients with hypertension. |
| 18 | Guideline for preoperative Assessment of coagulation profile | For interpretation of the coagulation profile |
| 19 | Guidelines on Pre-Operative Fasting for Elective Surgery | Fasting guidelines for surgery |
| 20 | Enhanced Recovery After Surgery for Vascular Patients undergoing Lower limb angioplasty with Distal Leg Wound | For patients undergoing angioplasty |
| 21 | Anaesthesia Protocol for Total/Unicompartmental Knee Arthroplasty (TKA/UKA) ERAS | ERAS for knee replacement operation |
| 22 | Anaesthesia Protocol for ERAS Spine | ERAS for spine operation |
| 23 | Breast Reconstruction ERAS protocol | ERAS for breast operation |
| 24 | Guideline for Enhanced Recovery after Surgery for Orthognathic Surgery | ERAS for Orthognathic operation |
| 25 | Enhanced Recovery After Surgery for Caesarean section | ERAS for cesarean section |
| 26 | Enhanced Recovery After Surgery for Open and Laparoscopic Liver Surgeries | ERAS for liver surgery. |
| 27 | Enhanced Recovery after Surgery for Benign Hysterectomy / Cystectomy / Myomectomy | ERAS for gynecology operation |
| 28 | Enhanced Recovery after Surgery for Oral Cavity Surgery with Free Flap Reconstruction | ERAS for free flap operation |
| 29 | Enhanced Recovery after Surgery for Colorectal surgery (Laparoscopic/Robotic/Open) | ERAS for colorectal operation |
| 30 | Guidelines on Perioperative Management of Diabetes Mellitus | For patients with diabetes |
| 31 | Guidelines on Preoperative Cardiac Assessment | For perioperative cardiac assessment |
| 32 | Anaesthesia Protocol for Hip Arthroplasty ERAS | ERAS for a hip operation |
| 33 | Guidelines on Anaemia | Preoperative anemia management |

| 34 | Guidelines on Perioperative Management of Adrenal Incidentaloma | For patients with adrenal incidentaloma |
| 35 | ACC/AHA Guideline on Perioperative Cardiovascular Evaluation and Management of Patients Undergoing Noncardiac Surgery: Executive Summary | For perioperative cardiac assessment |

**Supplementary Table 2:** Examples of what was keyed into the LLM-RAG model for the prompt, the clinical scenario (scenario 1) as well as the proposed correct answer.

| **Prompt for LLM-RAG model:** |
|---|
| You are the anesthesiologist seeing this patient in the preoperative clinic two weeks before the date of operation. The patients have already taken their routine preoperative investigations and the findings are listed within the clinical summary.<br>Your role is the evaluate the clinical summary and give the preoperative anesthesia instructions for the following patient targeted to your fellow medical colleagues. You are to follow strictly the department's guidelines.<br><br>Your instructions should consist of the following components:<br>1. Should the patient be seen by a Doctor or a Nurse - Doctor/Nurse<br>2. Fasting instructions - list instructions based on the number of hours before the time of the listed surgery<br>3. Suitability for preoperative carbohydrate loading - yes/no.<br>4. Medication instructions - name each medication and give the instructions for the day of the operation and days leading up to the operation as required.<br>5. Any instructions for the healthcare team - for example, preoperative blood group matching, arranging for preoperative dialysis, or standby post-operative high dependency/ICU beds.<br>6. Any preoperative optimization required for the patient - list what needs to be optimized.<br>7. Any need to delay the operation for further medical workup and preoperative optimization?<br>8. Any specific department protocols to follow for this patient - name as many as necessary, and give short reasoning for using these protocols.<br><br>Your instructions are the final instructions, do not give uncertain answers. If the medical condition is already optimized, there is no need to offer further optimization. If there are no relevant instructions in any of the above categories, leave it blank and write NA. |
| **Clinical scenario 1** |
| 38/Chinese/Female<br>Allergy to aspirin, paracetamol, penicillin - rashes and itchiness<br>ExSmoker - smoked 10 years ago / Occasional Drinker |

| | |
|---|---|
| LMP: last month Wt 94.7 Ht 166.3 BMI 34.2 BP 127/81 HR 88 SpO2 100% on RA<br><br>Coming in for BILATERAL REVISION FESS, REVISION SEPTOPLASTY, ADENOIDECTOMY AND BILATERAL INFERIOR TURBINOPLASTIES / SEVERE OSA ON CPAP<br><br>=== PAST MEDICAL HISTORY ===<br>1. Severe oSA on CPAP - AHI 58 - CPAP settings: AutoCPAP (4-15) cmH2O, without humidifier/Chinstrap<br>2. Right persistent Sinusitis<br>3. Allergic rhinitis<br>4. Adenoid hypertrophy<br>5. High BMI<br>6. Asthma - f/u GP, last seen 3 months ago for attack - on PRN ventolin - Does not use ventolin at all - No previous admissions/ intubations for asthma<br>7. Diabetes - HbA1C 9.4%, Last seen outpatient doctor >1 year ago.<br><br>No history of HTN/ HLD/ IHD/ CVA<br><br>=== SURGICAL HISTORY ===<br>Tonsillectomy > 10 years ago mild PONV<br><br>===Investigations===<br>Hb 13.0 TW 4 Plt 392<br>INR PT APTT normal<br>Na 134 K3.4 Cr 77 Glu 13<br>ECG NSR<br>CXR NAD<br><br>=== MEDICATIONS===<br>Ventolin PRN | |
| **Correct Answers:** | |
| Fasting instructions: | No food by mouth at least 6 hours before the operation, and no clear water up to 2 hours before the operation. |
| Suitability for preoperative carbohydrate loading: | No |
| Medication instructions | Bring ventolin on the morning of the operation |

| Instructions to the healthcare team | 1. List as morning case. |
| --- | --- |
| | 2. Do a hypocount on the morning of the operation. |
| | 3. Bring CPAP |
| | 4. Book Post-op HD |
| Preoperative optimization required | 1. Refer to the internal medicine perioperative team (IMPT) for diabetes optimization |
| Need to delay operation | Yes - Due to the need to improve diabetes control. |

**GPT4.0-RAG response**

1. Doctor
2. Fasting instructions:
   - Clear fluids up to 2 hours before surgery
   - No solid food or milk products after midnight on the day before surgery
3. No (due to high BMI and diabetes)
4. Medication instructions:
   - Ventolin: Continue as needed for asthma symptoms. Bring an inhaler to the hospital on the day of surgery.
   - Diabetes medications: Hold on the morning of surgery. Bring diabetes medications to the hospital.
   - CPAP: Continue use as usual until the time of surgery. Bring the CPAP machine to the hospital.
5. Instructions for the healthcare team:
   - Preoperative blood group matching: Not indicated unless specific surgical concerns arise.
   - Arrange for preoperative CPAP setup in the post-anesthesia care unit (PACU) or surgical ward.
   - Standby post-operative high dependency/ICU beds due to severe OSA and planned airway surgery.
6. Preoperative optimization required:
   - Optimize glycemic control, aiming for better HbA1c control. Consider endocrinology consultation if necessary.
   - Ensure asthma is well-controlled. If there are any concerns, consider pulmonary consultation.
7. Delay the operation for further medical workup and preoperative optimization:
   - Yes, consider delaying surgery to optimize diabetes control and ensure asthma management is optimized.
8. Specific department protocols to follow:
   - ERAS (Enhanced Recovery After Surgery) protocol for patients with OSA: This includes preoperative CPAP use, minimizing fasting periods, and early postoperative mobilization.
   - Asthma management protocol: Ensure asthma is well-controlled preoperatively, with access to rescue inhalers.

- Diabetes management protocol: Optimize glycemic control preoperatively and manage perioperative blood glucose levels.

**Supplementary Table 3:** Accuracy and hallucination rate of Preoperative instructions of the human-generated answers and the LLM and LLM-RAG answers.

| Models | Fasting Instructions | Preop Carbohydrate Loading | Medication Instruction | Instruction for Healthcare Team | Preoperative Optimization Required | Need to Delay Operation | Total correct | Hallucinations present |
|---|---|---|---|---|---|---|---|---|
| Human-Generated | 56/56 (100.0%) | 42/56 (75.0%) | 55/56 (98.2%) | 48/56 (85.7%) | 38/56 (67.9%) | 51/56 (91.1%) | 290/336 (86.3%) | |
| GPT3.5 | 14/14 (100.0%) | 10/14 (71.4%) | 11/14 (78.6%) | 8/14 (57.1%) | 6/14 (42.9%) | 10/14 (71.4%) | 59/84 (70.2%) | (1/84) 1.2% |
| GPT3.5-RAG | 13/14 (92.86%) | 9/14 (64.3%) | 11/14 (78.6%) | 11/14 (78.6%) | 8/14 (57.1%) | 12/14 (85.7%) | 64/84 (76.2%) | (3/84) 3.6% |
| GPT4.0 | 14/14 (100.0%) | 12/14 (85.7%) | 12/14 (85.7%) | 11/14 (78.6%) | 7/14 (50.0%) | 12/14 (85.7%) | 68/84 (80.1%) | (1/84) 1.2% |
| GPT4.0-RAG | 55/56 (98.2%) | 46/56 (82.1%) | 52/56 (92.9%) | 51/56 (91.1%) | 47/56 (83.9%) | 56/56 (100.0%) | 307/336 (91.4%) | (4/336) 1.2% |
| LLAMA2-7B | 14/14 (100.0%) | 10/14 (71.4%) | 7/14 (50.0%) | 7/14 (50.0%) | 4/14 (28.6%) | 10/14 (71.4%) | 52/84 (61.9%) | (10/84) 11.9% |
| LLAMA2-7B - RAG | 14/14 (100.0%) | 7/14 (50.0%) | 8/14 (57.1%) | 5/14 (35.7%) | 7/14 (50.0%) | 12/14 (85.7%) | 53/84 (63.1%) | (7/84) 8.3% |
| LLAMA2-13B | 13/14 (92.86%) | 8/14 (57.1%) | 8/14 (57.1%) | 7/14 (50.0%) | 5/14 (35.7%) | 9/14 (64.3%) | 50/84 (60.0%) | (13/84) 15.5% |
| LLAMA2-13B - RAG | 12/14 (85.7%) | 9/14 (64.3%) | 9/14 (64.3%) | 5/14 (35.7%) | 4/14 (28.6%) | 7/14 (50.0%) | 46/84 (54.8%) | (16/84) 19.1% |

**Supplementary Table 4:** Comparison between Human-generated and GPT4.0-RAG answers.

| Category | Cohen's h | 95% CI | Chi-square | p-value* |
|---|---|---|---|---|
| Fasting Instructions | 0.268 | (0.006, 0.530) | 0.009 | 0.924 |
| Preoperative Carbohydrate Loading | -0.175 | (-0.437, 0.087) | 0.477 | 0.490 |
| Medication Instructions | 0.273 | (0.011, 0.535) | 0.837 | 0.360 |
| Instructions for the Healthcare Team | -0.168 | (-0.430, 0.094) | 0.348 | 0.555 |
| Preoperative Optimization Required | -0.381 | (-0.643, -0.119) | 3.123 | 0.077 |
| Need to Delay the Operation | -0.606 | (-0.869, -0.345) | 3.349 | 0.067 |
| Overall | -0.199 | (-0.525, 0.177) | 0.260 | 0.610 |